\documentclass[10pt, a4paper]{article}
\usepackage[utf8]{inputenc} 
\usepackage[T1]{fontenc}    
\usepackage{hyperref}       
\usepackage{url}            
\usepackage{booktabs}       
\usepackage{amsfonts}       
\usepackage{nicefrac}       
\usepackage{microtype}      
\usepackage{lipsum}
\usepackage{graphicx}
\usepackage{comment}
\usepackage[english]{babel}
\graphicspath{ {./images/} }
\usepackage{diagbox}

\usepackage[]{lrec-coling2024} 

\usepackage{todonotes}
\usepackage{booktabs}

\title{The ParlaSent Multilingual Training Dataset for Sentiment Identification in Parliamentary Proceedings}

\name{Michal Mochtak, Peter Rupnik, Nikola Ljube\v{s}i\'{c}} 

\address{Dept. of Political Science, Dept. of Knowledge Technologies, Dept. of Knowledge Technologies\\
Radboud University, Jo\v{z}ef Stefan Institute, Jo\v{z}ef Stefan Institute \\
\texttt{michal.mochtak@ru.nl}, \texttt{peter.rupnik@ijs.si}, \texttt{nikola.ljubesic@ijs.si}\\}

\abstract{
The paper presents a new training dataset of sentences in 7 languages, manually annotated for sentiment, which are used in a series of experiments focused on training a robust sentiment identifier for parliamentary proceedings. The paper additionally introduces the first domain-specific multilingual transformer language model for political science applications, which was additionally pre-trained on 1.72 billion words from parliamentary proceedings of 27 European parliaments. We present experiments demonstrating how the additional pre-training on parliamentary data can significantly improve the model downstream performance, in our case, sentiment identification in parliamentary proceedings. We further show that our multilingual model performs very well on languages not seen during fine-tuning, and that additional fine-tuning data from other languages significantly improves the target parliament's results. The paper makes an important contribution to multiple disciplines inside the social sciences, and bridges them with computer science and computational linguistics. Lastly, the resulting fine-tuned language model sets up a more robust approach to sentiment analysis of political texts across languages, which allows scholars to study political sentiment from a comparative perspective using standardized tools and techniques.
 \\ \newline  }

\begin{document}

\maketitleabstract

\section{Introduction}
Emotions and sentiment in political discourse are deemed as crucial and influential as substantive policies promoted by the elected representatives~\cite{young2012affective}. Since the golden era of research on propaganda~\cite{lasswell_propaganda_1927,shils_cohesion_1948}, several scholars have demonstrated the growing role of emotions on affective polarization in politics with negative consequences for the stability of democratic institutions and social cohesion~\cite{garrett_implications_2014,iyengar_going_1995,mason_i_2015}. With the booming popularity of online media, sentiment analysis has become an indispensable tool for understanding the positions of viewers, customers, and voters~\cite{soler_twitter_2012}. It has allowed all sorts of entrepreneurs to know their target audience like never before~\cite{ceron_politics_2019}. Experts on political communication argue that the way we receive and process information plays an important role in political decision-making, shaping our judgment with strategic consequences both on the level of legislators and the masses~\cite{liu_appeal_2018}. Emotions and sentiment simply do play an essential role in political arenas, and politicians have been (ab)using them for decades.

Although there is a general agreement among political scientists that sentiment analysis represents a critical component for understanding political communication in general~\cite{young2012affective,flores2017,Tumasjan_Sprenger_Sandner_Welpe_2010}, the empirical applications outside the English-speaking world are still rare ~\cite{rauh2018,mohammad2021sentiment}. Moreover, many of the research applications in social sciences lag behind the latest methodological advancements grounded in computational linguistics. This is especially the case for studies analyzing political discourse in low-resourced languages, where the lack of out-of-the-box tools creates a huge barrier for social scientists to do such research in the first place~\cite{proksch2019,mochtak2020,rauh2018}. As a result, many of the applications still rely on dictionary-based methods, which tend to produce potentially skewed results~\cite{hardeniya_dictionary_2016, proksch2019} or approximate sentiment scores based on position-taking stances with relatively high-level generalization (e.g. roll-calls or voting behavior ~\cite{abercrombie_aye_2018}). Field-specific sentiment identifiers trained using machine learning algorithms are comparatively rare. Part of the reason is the fact that training machine learning models can be prohibitively expensive, especially when it comes to collecting, cleaning, and processing training data. However, recent development in the field of computational linguistic and natural language processing fueled by transformer-based deep learning models has lowered the bar for social scientists substantially. This development has additionally allowed for existing language models to be adapted to a specific domain by additionally pre-training the language model on non-annotated domain data~\cite{sung-etal-2019-pre}.

The paper presents annotated sentence-level datasets in seven European languages (Bosnian, Croatian, Czech, English, Serbian, Slovak, and Slovenian) sampled from parliamentary proceedings of seven European countries (Bosnia-Herzegovina, Croatia, Czech Republic, Serbia, Slovakia, Slovenia, and United Kingdom). The selection of proceedings is driven by the existing gap in low-resourced languages of Central and Eastern Europe and their relevance in a broader comparative perspective. The human-annotated datasets are used in a series of experiments focused on training sentiment identifiers intended for detecting sentiment in political discourse. Apart from methodological and experimental goals the paper has, it also can be read as summary guidelines for social scientists interested in training their own sentiment identifiers with similar scope. The paper is written with the intention of facilitating the process of collecting, cleaning, and processing textual data for political science research with realistic estimates for needed resources. When it comes to substantial findings, the paper shows that 1) additional pre-training of a language model on raw parliamentary data can significantly improve the model performance on the task of sentiment identification; 2) large 561-million-parameter multilingual models perform drastically better than those with half of the parameters; 3) multilingual models work very well also on unseen languages; and finally 4) even when the language-specific training data exist for the parliament proceedings one wants to process, a multilingual model trained on four times the size of the dataset from other languages improves the results on the target parliament significantly.

\section{Related work}

Despite the boom of computational methods in recent years has shown new ways to perform sentiment analysis with relatively high accuracy, political science is catching up relatively slowly. \citet{abercrombie_sentiment_2020} found that most of the automated applications focused on parliamentary debates and position-taking exist outside of the mainstream of political science, both a surprise and an opportunity for future research. Addressing the existing gap reflects upon the needs of empirical political science research, which recognizes that people tend to interact with politics through emotions \citep{masch_how_2020}. Recent research has found that political leaders are keen to use violent and populist rhetoric to connect 
with citizens on an emotional level~\cite{gerstle_negativity_2019,piazza_politician_2020,masch_how_2020}. As an effective campaigning strategy, populist parties in Europe use significantly more negative framing than their less populist counterparts simply because negative emotions work~\cite{widmann_how_2021}. They are often abused as highly conflicting drivers leading to affective polarization~\cite{druckman_affective_2021,iyengar_affect_2012}, negative partisanship~\cite{abramowitz_rise_2016}, group-based partisan competition ~\cite{mason_uncivil_2018}, and political sectarianism~\cite{finkel_political_2020}. If connected with the long-run emotional effects on the electorate, the impacts are disastrous. Partisan dehumanization, partisan antipathy, and acceptance of partisan violence are just a few examples of morphed competition infused with an emotionally laden identity fueling hostility, bias, and anger~\cite{webster_emotion_2022}. Understanding these mechanisms is highly important.


When it comes to actual applications focused on political domain, sentiment analysis can be most often found in research focused on classification~\cite{abercrombie_aye_2018,abercrombie_identifying_2018,akhmedova_co-operation_2018,bansal_power_2008,bonica_data-driven_2016} and dictionary-based sentiment extraction~\cite{honkela_five-dimensional_2014,onyimadu_towards_2013,mochtak_talking_2022,owen_exposure_2017,proksch2019}. The first stream of research uses different ML algorithms to develop models able to ``classify'' textual data into predefined categories (classes). These categories are either generated in an automated way based on known polarity traces, such as yes/no votes assigned to MPs’ speech acts~\cite{salah_machine_2015,abercrombie_aye_2018}, or are produced using traditional manual annotation with ground-truth labels~\cite{onyimadu_towards_2013,rauh2018}. A majority of these applications fall under the umbrella of supervised learning using a wide range of algorithms, from logistic regression to naïve Bayes, decision trees, nearest neighbor, or boosting. In recent years, many applications in computer science have been significantly tilted towards strategies using neural networks ranging from `vanilla' feed-forward networks to more complex architectures such as transformers pre-trained on raw non-annotated data~\cite{pipalia_comparative_2020}. In political science, however, dictionary-based strategies are still the dominant approach. They are traditionally focused on counting words with known sentiment affinity in raw text and generalizing their frequencies over the unit of analysis. Although sentiment dictionaries are deemed less accurate and may produce relatively crude estimates, their usage in political and social sciences is quite popular~\cite{mochtak_talking_2022,rinker_github_2017,abercrombie_sentiment_2020,proksch2019}. We see that as an opportunity for substantial improvement. The following sections present a new dataset for training a domain-specific sentiment identifier, which builds on a first-of-its-kind domain-specific transformer language model, additionally pre-trained on 1.72B domain-specific words from proceedings of 27 European parliaments. In a series of experiments, we then demonstrate how robust the approach is in various settings, proving its reliability in real-life applications. 


\section{Dataset construction}
\vspace{-0.3cm}
\subsection{Focus on sentences\label{sec:anno}}
\vspace{-0.2cm}
The datasets we compile and then use for training different prediction models focus on a sentence-level data and utilize sentence-centric approach for capturing sentiment polarity in text. The focus on sentences as the basic level of the analysis goes against the mainstream research strategies in social sciences which prefer either longer pieces of text (e.g. utterance of `speech segment' or whole documents~\cite{bansal_power_2008,thomas_get_2006}) or generally more coherent messages of shorter nature~\cite{Tumasjan_Sprenger_Sandner_Welpe_2010,flores2017}. However, this approach creates limitations when it comes to political debates in national parliaments, where speeches range from very short comments counting only a handful of sentences to long monologues with thousands of words. Moreover, as longer text may contain a multitude of sentiments, any annotation attempt must generalize them, introducing a complex coder bias that is embedded in any subsequent analysis. The sentence-centric approach attempts to refocus the attention on individual sentences capturing attitudes, emotions, and sentiment positions and use them as lower-level indices of sentiment polarity in a more complex political narrative. Although sentences cannot capture complex meanings as paragraphs or whole documents do, they usually carry coherent ideas with relevant sentiment affinity. This approach stems from a tradition of content analyses used by such projects as Comparative Manifesto~\cite{volkens_manifesto_2020}, the core-sentence approach in mobilization studies~\cite{hutter_politicising_2016}, or claims analysis in public policy research~\cite{koopmans_political_2006}.

Unlike most of the literature which approaches sentiment analysis in political discourse as a proxy for position-taking stances or as a scaling indicator~\cite{abercrombie_sentiment_2020, glavas_unsupervised_2017, proksch2019}, a general sentence-level classifier has a more holistic (and narrower) aim. Rather than focusing on a specific policy or issue area, the task is to assign a correct sentiment category to sentence-level data in political discourse with the highest possible accuracy. Only when a well-performing model exists, knowing whether a sentence is positive, negative, or neutral provides a myriad of possibilities for understanding the context of political concepts as well as their role in political discourse. Furthermore, sentences as lower semantic units can be aggregated to the level of paragraphs or whole documents, which is often impossible the other way around (document → sentences). Although sentences as the basic level of analysis are less frequent in political science research, existing applications include the validation of sentiment dictionaries~\cite{rauh2018}, ethos mining~\cite{duthie_deep_2018}, opinion mining~\cite{naderi_argumentation_2016}, or detection of sentiment carrying sentences~\cite{onyimadu_towards_2013}.

We base our experiments on data sampled from parliamentary proceedings which provide representative and often exhaustive evidence on political discourse in respective countries. In this context, parliamentary debates are considered to be a rich source of information on the position-taking strategies of politicians and one of the best sources of political discourse in general~\cite{lakoff_dont_2004}. As democracy thrives through debate, tracing it becomes essential to understanding politics, its development, and its consequences. In this context, essentially, all democratic parliaments hold a debate before a bill is passed. If public, the debate becomes evidence of how members of parliaments represent their voters and constituencies and their personal beliefs and interests~\cite{chilton_analysing_2004}. With all their flaws and shortcomings, parliamentary debates are an important aspect of political representation and an irreplaceable element of democratic systems. They connect voters with their representatives and show how responsive politicians are to people's wishes~\cite{powell_political_2004}. 
\vspace{-0.3cm}
\subsection{Background data}
\vspace{-0.2cm}
In order to compile datasets capturing political discourse, manually annotate them, and then use them for training the classification models for real world applications, we sampled sentences from seven corpora of parliamentary proceedings – Bosnia and Herzegovina~\citep{mochtak_bihcorp2022}\footnote{\url{https://doi.org/10.5281/zenodo.6517697}}, Croatia ~\citep{mochtak_crocorp2022}\footnote{\url{https://doi.org/10.5281/zenodo.6521372}}, Serbia ~\citep{mochtak_srbcorp2022}\footnote{\url{https://doi.org/10.5281/zenodo.6521648}}, Czechia ~\citep{erjavec_multilingual_2023} \footnote{\url{https://www.clarin.si/repository/xmlui/handle/11356/1486}}, Slovakia ~\citep{mochtak_svkcorp_2022} \footnote{\url{https://doi.org/10.5281/zenodo.7020474}}, Slovenia ~\citep{erjavec_multilingual_2023} \footnote{\url{https://www.clarin.si/repository/xmlui/handle/11356/1486}}, and United Kingdom ~\citep{erjavec_multilingual_2023} \footnote{\url{https://www.clarin.si/repository/xmlui/handle/11356/1486}}. The Bosnian corpus contains speeches collected on the federal level. Both chambers are included – House of Representatives (Predstavnički dom / Zastupnički dom) and House of Peoples (Dom naroda). The corpus covers the period from 1998 to 2018 and counts 127,713 speeches. The Czech corpus covers the period of 2013-2021 and counts 154,460 speeches from the Chamber of Deputies, the lower house of the parliament (Poslanecká sněmovna). The Croatian corpus of parliamentary debates covers debates in the Croatian parliament (Sabor) from 2003 to 2020 and counts 481,508 speeches. The Serbian corpus contains 321,103 speeches from the National Assembly of Serbia (Skupština) over the period of 1997 to 2020. The Slovenian corpus covers the period of 2000-2022 and counts 311,354 speeches from the National Assembly (Državni zbor), the lower house of the parliament. The Slovak corpus contains speeches from the period of 1994-2020 from the National Council of the Slovak Republic (Národná rada) and counts 375,024 speeches. Finally, the corpus from British Parliament covers speeches from both the House of Commons and the House of Lords from the period of 2015-2021 counting 552,103 speeches.
\vspace{-0.3cm}
\subsection{Data sampling}
\vspace{-0.2cm}
Speeches were segmented into sentences and words using the CLASSLA-Stanza~\cite{ljubesic-dobrovoljc-2019-neural,terčon2023classlastanza} and Trankit ~\cite{nguyen_trankit_2021} pipelines with tokenizers available for Czech, Croatian, Serbian, Slovak, Slovene, and English languages. This step was necessary in order to extract individual sentences as the basic unit of our analysis. In the next step, we filtered out only sentences presented by actual speakers, excluding moderators of the parliamentary sessions. Sentences are preserved within their country-defined pools with the exception of Bosnia and Herzegovina, Croatia, and Serbia which were merged together as representatives of one language family (i.e. the corpora were treated as one sampling pool). We are, from now on, referring to this pool as BCS (Bosnian-Croatian-Serbian)\footnote{The main reason for keeping these three parliaments in one pool is 
previous work on annotating sentiment in parliamentary proceedings~\cite{mochtak2022parlasent} 
which consisted of 2,600 instances jointly sampled from the three underlying parliaments.}. As we want to sample what can be understood as ``average sentences'', we further subset each sentence-based corpus to only sentences having the number of tokens within the first and third frequency quartile (i.e. being within the interquartile range) of the original corpora. Having the set of ``average sentences'', we used common sentiment lexicons available for each of the languages ~\cite{glavas_semi-supervised_2012, chen_building_2014, chen2016false}, and applied them as seed words for sampling sentences for manual annotation. These seed words are used for stratified random sampling which gives us 867 sentences with negative seed word(s), 867 sentences with positive seed word(s), and 866 sentences with neither positive nor negative seed words (supposedly having neutral sentiment) per sampling pool. We sample 2,600 sentences in total for manual annotation per corpus. Under this setting, the sentences inherit all metadata information of their parent documents.

We further sample two random datasets, one from the BCS collection of parliaments, another from the English parliament, not applying the sentiment seed list, but rather aiming at a random selection of sentences, still satisfying the criterion of the sentence length falling into the interquartile range. The primary use case for these two datasets is testing various sentiment identification approaches, therefore we wanted for their sentiment distribution to follow the one occurring in the underlying parliaments. The sampling pipeline is identical to seed-based datasets but without the seed words component. For our experiments, we again sample 2,600 average-length sentences cleaned off of entries from proceedings' moderators (see above). From now on, we refer to these two datasets as the BCS-test and the EN-test sets.

\subsection{Annotation schema\label{sec:anno}}

The annotation schema for labelling sentence-level data was adopted from  \citet{batanovic_versatile_2020} who propose a six-item scale for annotation of sentiment polarity in a short text. The schema was originally developed and applied to SentiComments.SR, a corpus of movie comments in Serbian and is particularly suitable for low-resourced languages. The annotation schema contains six sentiment labels \cite[p.~6]{batanovic_versatile_2020}:

\begin{itemize}
    \item +1 (\texttt{Positive} in our dataset) for sentences that are entirely or predominantly positive
    \item –1 (\texttt{Negative} in our dataset) for sentences that are entirely or predominantly negative
    \item +M (\texttt{M\_Positive} in our dataset) for sentences that convey an ambiguous sentiment or a mixture of sentiments, but lean more towards the positive sentiment in a positive-negative classification
    \item –M (\texttt{M\_Negative} in our dataset) for sentences that convey an ambiguous sentiment or a mixture of sentiments, but lean 	more towards the negative sentiment in a positive-negative classification
    \item +NS (\texttt{P\_Neutral} in our dataset) for sentences that only contain non-sentiment-related statements, but still lean more towards the positive sentiment in a positive-negative classification
    \item –NS (\texttt{N\_Neutral} in our dataset) for sentences that only contain non-sentiment-related statements, but still lean more towards the negative sentiment in a positive-negative classification

\end{itemize}

The different naming convention we have applied in our dataset serves primarily practical purposes: obtaining the 3-way classification by taking under consideration only the second part of the string (if an underscore is present).

The benefit of the whole annotation logic is that it was designed with versatility in mind allowing reducing the sentiment label set in subsequent processing if needed. That includes various reductions considering polarity categorization, subjective/objective categorization, or change of the number of categories. This is important for various empirical tests we perform in the following sections.
\vspace{-0.3cm}
\subsection{Data annotation\label{sec:dataanno}}
\vspace{-0.2cm}
Data were annotated in multiple iterations. Each seed-based dataset was annotated by two annotators, both being native speakers or having an excellent command of the language to be annotated. Annotation was done via a custom-built, locally-hosted online app using \textit{prodigy} with consistent logging allowing monitoring of the annotation process systematically ~\cite{prodigy2018}. Each annotator went through approximately ten hours of training before the actual annotation. The annotation was done iteratively in several rounds, with automated oversight of the coding process in order to minimize any systematic disagreement. Trained annotators were able to annotate up to 75 sentences per hour on average, resulting in 35 person-hours per annotator, per dataset (feedback and reconciliation not included). When it comes to BCS and English test sets, data were annotated by only one highly trained annotator. Similarly to previous setting, annotation of both test sets were followed by quality control procedures adjusted for just one annotator (e.g., consistency monitoring; pace; consultations).

Despite the relatively smooth annotation process across the datasets, the inter-annotator agreement (IAA) measured using Krippendorff's alpha (KA) has not reached high values. This is consistent across datasets supporting the argument that sentiment perception is a highly subjective matter~\cite{bermingham_using_2011, mozetic_multilingual_2016} and, despite the effort to eliminate hard disagreements, the results often reflect upon a hard call the annotators had to make. Monitoring and reconciliation of the disagreements further showed that most of the disagreements are not substantially wrong and can be considered relevant under the provided context (i.e. shorter text snippets). The summary of Krippendorff's alpha and agreement rates (accuracy) across datasets and annotation schemas is presented in Table~\ref{tab:distr1}.\footnote{We do not report KA for the BCS and English test sets here as only one annotator performed the annotations.}

\begin{table*}
\begin{center}
\begin{tabular}{|l|r|r||r|r|}
\hline
Dataset & ACC (6 classes) & KA (6 classes) & ACC (3 classes) & KA (3 classes)   \\
\hline
BCS & 62.0\% & 0.502 & 77.7\% & 0.639\\
CZ & 68.1\% & 0.531 & 77.4\% & 0.612\\
SK & 63.4\% & 0.506 & 75.4\% & 0.607\\
SL & 64.1\% & 0.502 & 73.7\% & 0.531\\
EN & 66.0\% & 0.543 & 78.4\% & 0.656\\
\hline
 
\end{tabular}

\end{center}
\caption{\label{tab:distr1} Krippendorff's alpha (KA) and agreement rates (ACC) across datasets for 6-fold and 3-fold annotation schemas.}
\end{table*}

The final distributions of the 
three-class  labels after reconciliation across datasets are presented in 
Table~\ref{tab:distr3}.
The reconciliation was done by the original annotators after the main annotation round was finished. As the reconciliation was done by mutual agreement, the whole process was administrated online to avoid any immediate peer pressure. Annotators were first asked to mark annotations for which they do not have problem to agree with their colleague, indicating their reconciliation position without the need for actual deliberation (i.e., discussion). This approach helped to eliminate disagreements which can be considered easy. Annotators then met in person or online and discussed instances which they could not agree on in the first round. Each dataset contained a handful of hard disagreements that were not possible to reconcile without an external reviewer (one of the authors of this paper). The presented distributions show that the negative class is often the most populous category. This observation, however, does not entirely hold true in the Slovene dataset and English test set (see Table~\ref{tab:distr3}). We theorize that this might indicate potential nuances in political nature of different parliaments and an existence of different patterns in their political culture. Neutral sentiment appears to be more dominant there.

\begin{table}[hbt!]
\begin{center}
\begin{tabular}{|l|r|r|r|}
\hline
Dataset & Negative & Neutral & Positive    \\
\hline
all & 8232 & 6691 & 3277 \\
BCS & 1314 & 773 & 513 \\
CZ & 1398 & 866 & 336 \\
SK & 1253 & 895 & 452  \\
SL & 1010 & 1409 & 181  \\
EN & 1269 & 680 & 651  \\
BCS-test & 1147 & 1006 & 447 \\
EN-test & 841 & 1062 & 697 \\
\hline

\end{tabular}

\end{center}
\caption{\label{tab:distr3}Distribution of the three-class labels across datasets.}
\end{table}

\vspace{-0.2cm}
\subsection{Dataset encoding\label{sec:encoding}}

The final dataset is encoded as a JSONL document, each line in the dataset representing a JSON object. The dataset is encoded in seven files, five files representing the 2,600 training instances per language (group) and two files representing our two BCS and English test sets.

Each of the training datasets contains the following metadata:

\begin{itemize}
    \item \texttt{sentence} that is annotated.
    \item \texttt{country} of origin of the sentence
    \item annotation of \texttt{annotator1} with one of the labels from the annotation schema presented in Section~\ref{sec:anno}
    \item annotation of \texttt{annotator2} following the same annotation schema
    \item annotation given during \texttt{reconciliation} of hard disagreements
    \item the three-way \texttt{label} (positive, negative, neutral) where +NS and -NS labels are mapped to the neutral class
    \item the \texttt{document\_id} the sentence comes from
    \item the \texttt{sentence\_id} of the sentence
    \item the \texttt{term} inside which the speech was given
    \item the \texttt{date} the speech was given
    \item the \texttt{name}, \texttt{party},  \texttt{gender}, \texttt{birth\_year} of the speaker
    \item the \texttt{split} the instance has been assigned to (training set containing 2054 instances, development set 180 instances, and testing set 366 instances)    
    \item the \texttt{ruling} status of the party while the speech was given (opposition or coalition)
\end{itemize}


The EN-test and BCS-test sets differ in their structure minimally, containing only the \texttt{annotator1} attribute, and missing the \texttt{annotator2} and \texttt{reconciliation} attributes, as single annotator performed the annotation of test sets. Furthermore, there is no \texttt{split} attribute as the purpose of these datasets is testing various algorithms, while the training datasets can also be used for language-specific experiments, therefore requiring the train-dev-test split.

The dataset is made available through the CLARIN.SI repository at 
\url{http://hdl.handle.net/11356/1585}
and is available under the CC-BY-SA 4.0 license.
\vspace{-0.3cm}
\section{Experiments}
\vspace{-0.2cm}
In this section, we present our experiments through which we aim to answer the following three research questions:

Q1: Does our newly released XLM-R-parla model, which is an XLM-R-large model additionally pre-trained on 1.72 billion words of parliamentary proceedings, model the sentiment phenomenon better than the original XLM-R models?

Q2: How well does our model work on languages not seen during fine-tuning?

Q3: If one wants to process data from some parliament that is covered in the ParlaSent (or any other) dataset, is it advisable to train a language-specific (and parliament-specific) model, or rather train a multilingual model containing the whole ParlaSent dataset?

To make the most out of our rather complex 6-level annotation schema, we set-up all our experiments as regression tasks, the six levels being modified into integer values from 0 to 5. For evaluation, we use primarily the $R^2$ score, which quantifies the proportion of the prediction variance that can be explained through gold labels, due to its sensitivity to differences in system performance. We also report mean average error (MAE) as a simple-to-understand metric in terms of an average error per instance, knowing that we are predicting values on a scale from 0 to 5, maximum MAE thereby being 5, and acceptable error for most use cases being somewhere below 1.
\vspace{-0.3cm}
\subsection{The XLM-R-parla model}
\vspace{-0.2cm}
In this subsection we are answering our first question -- whether our newly released model XLM-R-parla\footnote{\url{https://huggingface.co/classla/xlm-r-parla}} 
performs better than the vanilla XLM-R models of size base and large~\cite{conneau2019unsupervised}.

The XLM-R-parla model is based on the XLM-R-large model\footnote{\url{https://huggingface.co/xlm-roberta-large}} due to our preliminary experiments showing that XLM-R-large models outperform XLM-R-base models on this task. The XLM-R-parla model was additionally pre-trained for only 8 thousand steps with a batch size of 1024 sequences of 512 tokens of text. The model was pre-trained on a merge of the ParlaMint 3.0 dataset~\citep{Parlamint30} and the EuroParl dataset~\citep{europarl}, together covering 30 languages, with 1,717,113,737 words of running text. Important to mention is that our pre-training dataset consists of all the languages contained inside the ParlaSent dataset. 

Our hypothesis for this question is that the additional adaptation of the XLM-R-large model to texts as they occur in parliamentary proceedings will significantly improve our predictions of sentiment in parliamentary proceedings.

We fine-tune both the XLM-R-base and XLM-R-large, as well as the XLM-R-parla model, with the same hyperparameter settings, that have proven in preliminary experiments to perform well for our task: learning rate of 8e-6, batch size of 32, and 4 epochs over our whole training dataset (2,600 instances per each of the five parliament( pool)s, 13,000 instances in total). We test each model separately on the BCS and the English test set, each comprising of 2,600 instances. We compare these three models with a dummy baseline returning always the mean value of the training data. We perform five runs for each set-up and report the mean result.

\begin{table}
\centering
\begin{tabular}{|l|rr|rr|}
\hline
& \multicolumn{2}{c|}{$R^2$} & \multicolumn{2}{c|}{$MAE$}\\
model \textbackslash{} test & BCS & en & BCS & EN \\
\hline
Dummy & -0.012 & -0.165 &  1.522 & 1.645 \\
XLM-R-base & 0.500 &	0.561 & 
0.868 & 	0.808 \\
XLM-R-large & 0.605 &	0.653 &	0.706	& 0.694 \\
XLM-R-parla & 0.615 & 0.672 & 0.705 & 	0.675 \\
\hline
\end{tabular}
\caption{\label{tab:q1}Results of the first research question comparing the additionally-pretrained XLM-R-parla model with the vanilla XLM-R models and a random baseline. \vspace{-0.5cm}}
\end{table}

The results in Table~\ref{tab:q1} show that the mean dummy, as expected, gives a $R^2$ value of around 0, while the MAE is around 1.5, which means that, if we always predicted a mean value from our training data, on average, we would be ``only'' 1.5 points away from the correct value. This result represents the baseline result any reasonable system should improve over. The XLM-R-base model does exactly that, lowering the MAE to between 0.81 and 0.87, depending on the test set. 

The XLM-R large model, identical to our preliminary experiments, drastically improves over the base-sized model, which simply shows that the task at our hands is a rather complex one and that the extra capacity delivers around 0.1 points better results in $R^2$ (scale 0-1), which can be called a drastic improvement. The MAE score, much less sensitive to changes in the quality of predictions, still shows an error lower on average of 0.1 to 0.15 points (scale 0-5).

Once we compare the original XLM-R-large model with the additionally pre-trained XLM-R-parla model, we can observe that the additional pre-training has paid off, with minor, but consistent improvements on all metrics.

As expected, all systems perform better on the English test set due to much more English data  seen during pre-training than that of Bosnian, Croatian, or Serbian.

We can conclude by offering an answer to the first research question – the additional pre-training of a multilingual model on parliamentary data does improve the performance on our task.
\vspace{-0.3cm}
\subsection{Performance on unseen languages}
\vspace{-0.2cm}
Here, we are answering our second question – how is the performance of our best-performing model XLM-R-parla on a language that the model has not seen during fine-tuning?

Our initial hypothesis is that there would be a minor impact on whether the model was fine-tuned on the testing language or not.

We perform two ablation experiments, in one skipping BCS data from the training dataset, and in the other skipping English data, and we evaluate both models on the BCS and the English test sets. Therefore, in the two additional experiments, we do not train on 13,000 but on 10,400 instances. We keep the same hyperparameter values as with the initial XLM-R-parla experiment described in the previous subsection.

\begin{table*}
\centering
\begin{tabular}{|l|rr|rr|}
\hline
 & \multicolumn{2}{c|}{$R^2$} & \multicolumn{2}{c|}{MAE} \\
training set  & BCS & en & BCS & en \\
 \hline
ParlaSent & 0.615 & 0.672 & 0.705 & 	0.675 \\
ParlaSent $\textbackslash \{BCS\}$ & 0.630 &	0.659 &0.727 & 0.704 \\
ParlaSent $\textbackslash \{EN\}$  & 0.596 &  0.655 & 0.728 &	0.756 \\
\hline
\end{tabular}
\caption{\label{tab:q2}Experiments on removing the BCS and English data from the training data, evaluating on the BCS and English test data, to check for performance on languages not seen during fine-tuning. \vspace{-0.3cm}}
\end{table*}

In Table~\ref{tab:q2}, reporting the results of these experiments, to our surprise, we cannot observe any obvious pattern regarding the performance on the language that has been removed from the training data. On the BCS example the model performs even better on the BCS test set regarding the $R^2$ evaluation metric and worse on the EN dataset. If we look at the MAE metric, the results are slightly more expected, the model performing worse on both test sets, the drop being still more significant on the English test set. The inconsistency between the two metrics on the BCS test set is very likely due to $R^2$ penalizing outliers more harshly than MAE does.

In the experiment where the English training data is removed, the results are a little more consistent, with performance on both test sets being similarly worse, the English test set getting an extra hit on the MAE metric but not on the $R^2$ metric.

Overall, we cannot observe a hit on the performance of the models if the testing language gets removed from the training data to a greater extent than what is observed on the other test set. Therefore, we can conclude that the performance drops due to less training data, not due to the target language not being present in the training data.

\vspace{-0.3cm}
\subsection{Monolingual vs. multilingual training}\vspace{-0.2cm}
In this subsection, we answer our third research question – whether target language performance is better if the model is trained on that language only, or rather if it is trained on all five parliament( group)s.

We hypothesize that results might be evened out. On the monolingual side, there is a drastic similarity between the training and the test data, not just due to the same language used, but also due to data coming from the same parliament, each parliament surely having many specific features a system might use to predict sentiment. On the multilingual side, the argument relies on five times more training data than in the monolingual setting.

In this set of experiments, we train and evaluate only on the training datasets, which have a train-dev-test split, as already reported in Section~\ref{sec:encoding}. In the case of the monolingual setting, we train the model only on the 2054 training instances for the specific parliament, while in the case of the multilingual setting, we train on five times that amount of data, i.e., 10,270 instances. We always evaluate on the test portion of the training dataset of the target parliament. We keep our hyperparameters the same as before, with the difference that monolingual models are fine-tuned for ten epochs due to less data available for fine-tuning.

\begin{table}
\centering
\begin{tabular}{|l|rr|rr|}
\hline
& \multicolumn{2}{c|}{$R^2$} & \multicolumn{2}{c|}{MAE} \\
language & mono & multi & mono & multi \\
\hline
bcs & 0.699 & 0.737 & 0.644 & 0.572 \\
cz & 0.564 & 0.560 & 0.706 & 0.665 \\
en & 0.707 & 0.741 & 0.652 & 0.599 \\
sk & 0.646 & 0.681 & 0.665 & 0.593 \\
sl & 0.512 & 0.520 & 0.708 & 0.667 \\
\hline
\end{tabular}
\caption{\label{tab:q3}Experiments on comparing performance when training on the target language vs. training on all available languages. \vspace{-0.5cm}}
\end{table}

The results in Table~\ref{tab:q3} show that only in Czech there seems to be a similar performance of the monolingual and the multilingual models, while in all remaining parliaments, there is a consistent benefit of training on all available languages. Given these results, we can conclude that if one wanted to annotate the sentiment in a specific parliament for which there is training data available, better results might still be obtained with additional data, written in different languages and coming from different parliaments.

This result also shows that our annotation guidelines were detailed enough for the annotators in the different languages to have comparable annotation criteria, thereby rendering annotations in different languages useful to each other.


\vspace{-0.3cm}
\section{Conclusion}
\vspace{-0.2cm}
In this paper, we have presented a new dataset, consisting of sentences coming from seven different parliaments, manually annotated with a six-level schema for sentiment. This is the first of such datasets available for parliamentary proceedings' data. We show the inter-annotator agreement to be reasonably high for such an endeavor. We share 2,600 instances per parliament (group), the Bosnian, Croatian, and Serbian parliaments forming a single BCS parliament group, the remaining four parliaments being those of the Czech Republic, United Kingdom, Slovakia, and Slovenia. Aside from these five training datasets, we also share two additional test sets, one from the BCS group and another from the United Kingdom. The data are shared via the CLARIN.SI repository\footnote{\url{http://hdl.handle.net/11356/1868}}.

In our experiments, we answer three main research questions. The first question relates to whether additional pre-training of a transformer model on raw parliamentary data would improve the performance on the task, which proves to be correct, and therefore, we also present the new XLM-R-parla model\footnote{\url{https://huggingface.co/classla/xlm-r-parla}}, especially suited for parliamentary text processing. Whether the model is more potent in the processing of political texts in general, will have to be tested in follow-up work.

The second question we tackle is how well we can expect the final, fine-tuned model, named XLM-R-ParlaSent\footnote{\url{https://huggingface.co/classla/xlm-r-parlasent}},
to perform on languages and parliaments not seen during fine-tuning. We show that the model is very robust to unseen languages and parliaments with no or minor drop in performance. The limitation of this insight is that the languages and parliaments we check this presumption on are linguistically and traditionally rather related to the remaining languages and parliaments in the training data, so caution is advised if more distant languages or parliaments were to be annotated with the XLM-R-ParlaSent model.

The third question relates to whether a model performs better on a specific language and parliament if it is trained on that parliament's data only, or whether the additional, four times larger dataset, coming from different languages and parliaments, is more beneficial. We show that the multilingual multi-parliamentary approach performs better, which is a direct signal that our annotations are not of high quality inside parliaments only, as measured via the inter-annotator agreement, but also between parliaments and languages. 


We consider this work to be a very important step in setting up a more robust approach to sentiment analysis of political texts beyond sentiment lexicon approaches, which will find many applications in the downstream research of political and parliamentary communications. It is part of a more general effort to democratize the latest advancements in natural language processing and their relevance in humanities and social sciences.

\section{Acknowledgments}
The research presented in this paper was conducted within the research project ``Basic Research for the Development of Spoken Language Resources and Speech Technologies for the Slovenian Language'' (J7-4642), the research project ``Embeddings-based techniques for Media Monitoring Applications'' (L2-50070, co-funded by the Kliping d.o.o. agency) and within the research programme ``Language resources and technologies for Slovene'' (P6-0411), all funded by the Slovenian Research and Innovation Agency (ARIS).

\bibliographystyle{lrec-coling2024-natbib}
\bibliography{paper}

\bibliographystylelanguageresource{lrec-coling2024-natbib}
\bibliographylanguageresource{languageresource}

\end{document}